\documentclass[english]{llncs}
\usepackage[LGR,T1]{fontenc}
\usepackage[latin9]{inputenc}
\usepackage{geometry}
\usepackage{color}
\usepackage{amsmath}
\usepackage{amssymb}
\usepackage{graphicx}

\makeatletter

\DeclareRobustCommand{\greektext}{%
  \fontencoding{LGR}\selectfont\def\encodingdefault{LGR}}
\DeclareRobustCommand{\textgreek}[1]{\leavevmode{\greektext #1}}
\ProvideTextCommand{\~}{LGR}[1]{\char126#1}

\makeatother

\usepackage{babel}
\begin{document}
\global\long\def\ve#1{\mathbf{#1}}%

\title{On Learning and Learned Representation with Dynamic Routing in Capsule
Networks}
\author{\author{Ancheng Lin\inst{1} \and Jun Li\inst{2}\and Zhenyuan Ma\inst{3,}\thanks{Corresponding author, email: mazy@gpnu.edu.cn} \authorrunning{A. Lin et al.}}\institute{School of Computer Sciences, Guangdong Polytechnic Normal University, Guangzhou, China   \email{cenbylin@163.com}\\ \and School of Software and Centre for Artificial Intelligence, Faculy of Engineering and Information Technology, University of Technology Sydney, POBox 123, Broadway, NSW 2007, Australia \email{Jun.Li@uts.edu.au}\\ \and School of Mathematics and System Sciences, Guangdong Polytechnic Normal University, Guangzhou, China  \email{mazy@gpnu.edu.cn}}}
\maketitle
\begin{abstract}
Capsule Networks (CapsNet) are recently proposed multi-stage computational
models specialized for entity representation and discovery in image
data. CapsNet employs iterative routing that shapes how the information
cascades through different levels of interpretations. In this work,
we investigate i) how the routing affects the CapsNet model fitting,
ii) how the representation by capsules helps discover global structures
in data distribution and iii) how learned data representation adapts
and generalizes to new tasks. Our investigation shows: i) routing
operation determines the certainty with which one layer of capsules
pass information to the layer above, and the appropriate level of
certainty is related to the model fitness, ii) in a designed experiment
using data with a known 2D structure, capsule representations allow
more meaningful 2D manifold embedding than neurons in a standard CNN
do and iii) compared to neurons of standard CNN, capsules of successive
layers are less coupled and more adaptive to new data distribution.

\end{abstract}

\section{Introduction}

Deep neural networks have achieved great success in image and video
processing tasks. Capsule Net (CapsNet) \cite{Sabour2017} is a recently
proposed architecture that represents an alternative arrangement of
multiple stage processing of image data. Essentially, CapsNet differs
from the traditional deep neural networks in i) in each stage, the
atomic units of information are vectors rather than scalar values\footnote{To be more specific, the ``atomic units'' refer to the basic random
variables that we are concerned with. In practical image/video analytic
tasks, this concept of lower ``convolutional'' layers correspond
to an element of a channel at a particular image location. }, and ii) the output of a processing stage no longer contributes equally
to the computation of its successive stage. One can intuitively understand
the changes as introducing structures in the information and the information
no longer flows homogeneously through the processing pipeline.

We present in this work our investigation on three aspects of CapsNet,
answering three fundamental questions about the effectiveness of the
new architecture in terms of visual analytics, namely model fitting,
representation learning and generalization: 
\begin{itemize}
\item Model fitting: how the routing process affects the training of the
network. Routing introduces additional dynamics in the information
flow through a network. Different from conventional neural networks,
where cross-layer connections are determined completely by network
parameters. The connections between two successive layers of capsules
in a CapsNet are computed at run-time and vary with individual data
samples. Our investigation shows routing determines the (un-)certainty
in the information pathway through layers of capsules. The appropriate
level of certainty is closely interwoven with the model fitness to
data.
\item Representation learning: one motivation of CapsNet is that capsule
representation can correspond to interpretable attributes of images,
such as the style of writing in the case of hand-written digits. In
this work, we test trained models on image data on a known 2D manifold
spanned by geometric transformations. The ground-truth data manifold
allows us to quantitatively assess the data representations in terms
of meaningful structure discovery. We found that compared to standard
neural networks, CapsNet captured more faithfully the global manifold
structure in the image data. 
\item Representation generalization:  if CapsNet could recover parse tree-like
structures for images \cite{Sabour2017}, the capsules would correspond
to entities at different levels of interpretation. One can then expect
the intermediate-level capsules to correspond to cognition ingredients
that could be re-adapted to new tasks or contexts. In our comparative
study, CapsNet generated mid-level data representations more adaptable
to new tasks than conventional neural networks did, which supported
the claim above. 
\end{itemize}
In the remaining parts of this paper, we review necessary background
in Section \ref{sec:rev}. \mbox{Section \ref{sec:tst}} presents
the main findings of the research in three aspects. Section \ref{sec:con}
concludes the paper.\vspace{-5pt}

\section{Background Review\label{sec:rev}}

Research in deep neural networks (DNN) have enjoyed rapid growth in
recent years \cite{Lecun2015}. DNN-powered learning models have represented
the state-of-the-art in a wide range of application areas \cite{He2016,Krizhevsky2012,Rafi2016,Altwaijry2016,Garbade2016}.
On the other hand, fundamental challenges remain in artificial neural
network-based vision systems. The training process is complex and
expensive in terms of both computation and training data \cite{Lecun2015,Jia2017}.
The learning is task-oriented and end-to-end, where the understanding
of intermediate data representation is incomplete and the decision
making is obscure \cite{Koh2017}. The insufficiently understood model
may be prone to peculiar failures or attacks \cite{Sharif2016,Papernot2017}.
Generalization and reliability of an existing DNN beyond the training
domain can also be problematic \cite{Yosinski2014}.

CapsNet \cite{Sabour2017} represents an alternative visual information
processing mechanism that addressing some abovementioned issues. The
neurons are divided into small groups in each network layer, known
as \emph{capsules}. The capsules correspond to concepts in different
levels of abstraction during the process of parsing visual information.
The cross-layer association and the activation status of the capsules
represent semantic analysis of the image data.  Recently, CapsNet
has undergone some developments such as Matrix Capsules \cite{Hinton2018}
and has been employed in new application domains such as text classification
\cite{Zhao2018}. 

The investigation into CapsNet in this work is mostly related to three
topics of research in DNN and broadly machine learning: model training,
data representation and knowledge transfer. Arguably, efficient training
methodology plays the midwife for real-world success of DNN \cite{Hinton2006}.
Rich techniques have been proposed to address different challenges
in various DNN structures, including randomly disturbing cross-layer
connections \cite{Srivastava2014}, introducing special gate units
to keep long-term memory \cite{Hochreiter1997,Kataoka2016} and exploiting
computationally affordable structures in the gradients during optimization
\cite{Kingma2014}. The iterative routing in CapsNet is a newly introduced
technique, where its role in model training demands full investigation.

Data representation is one of the key elements of successful analysis
\cite{Hastie2009}. It is well-known that the learned convolutional
neurons in the lower layers of deep networks resemble the primary
biological vision processing in discovering low-level features in
images \cite{Reza2016}. On the other hand, the roles of intermediate
or high-layer neurons in DNNs are not well understood \cite{Karpathy2015,Koh2017}.
 To address this issue, CapsNet was proposed and has shown the promise
of unveiling the meaningful data structures in image populations.
In this work, we perform a comprehensive study on CapsNet and propose
a systematic quantitative evaluation protocol.

One advantage of general AI is its supreme adaptability.  Tremendous
research focus has been placed in transfer learning \cite{Pan2010}.
In particular, pioneering investigation has revealed characteristics
of DNN layers under transfer tasks \cite{Yosinski2014}. The abovementioned
meaningful capsule data representation indicates capsules are conducive
to knowledge transfer, which is supported by experiments in Section
\ref{sec:tst}.

In following discussion, we will frequently consider the routing
between adjacent layers of capsules. Below we provide a brief review
of routing; readers can refer to \cite{Sabour2017} for more details.
The pre-activation (total-input) of a capsule $j$ is a vector $\ve s_{j}=\sum_{i}c_{ij}\hat{\ve u}_{j|i}$,
where $\hat{\ve u}_{j|i}$ is the prediction of $j$ by a low-layer
capsule $i$ and $c_{ij}$ is the association coefficient between
$i$ and $j$. Routing determines the association coefficients by
iterations of,
\begin{itemize}
\item accumulating alignment between activated capsule $j$, $\ve v_{j}=\mathrm{a}(\ve s_{j})$,
and $\hat{\ve u}_{j|i}$ , where $\mathrm{a}(\cdot)$ is a non-linear
activation and $\hat{\ve u}_{j|i}$ is a ``prediction vector'' from
a lower capsule i, \vspace{-5pt}
\begin{align}
b_{ij} & \leftarrow b_{ij}+\bigl\langle\ve v_{j},\hat{\ve u}_{j|i}\bigr\rangle\label{eq:bij}
\end{align}
\item updating $c_{ij}$ using accumulated alignments, \vspace{-5pt}
\begin{align}
c_{ij} & \leftarrow\mathrm{Softmax}(b_{ij})=\frac{e^{b_{ij}}}{\sum_{j}e^{b_{ij}}}\label{eq:cij}
\end{align}
Note $\ve v_{j}$ in (\ref{eq:bij}) relies on coefficients $\{c_{ij}\}$
obtained in (\ref{eq:cij}). 
\end{itemize}

\section{Experiments\label{sec:tst}}

\subsection{Model fitting\label{subsec:model-fitting}}

This section presents our experiment results and analysis of structural
and operational factors that affects the model training process of
CapsNet. Multi-layer neural networks are powerful generic function
approximation models, while the new family of CapsNet\cite{Sabour2017}
introduce extra versatility via routing in data representation which
are particularly effective in extracting the semantic hierarchy embedded
in sensory data. Nevertheless, a data model can only realize its potential
if there is an effective way of fitting the model to data.  So it
is natural to ask what CapsNet has provided considering the trade-off
between model capability and training complexity. In particular, data
representation routing is realized as Expectation-Maximization (EM)
inference on the association between two layers of capsule units.
We test and analyze how the EM operations affect the model training
and performance. 

\begin{figure}
\begin{centering}
\includegraphics[width=0.9\columnwidth]{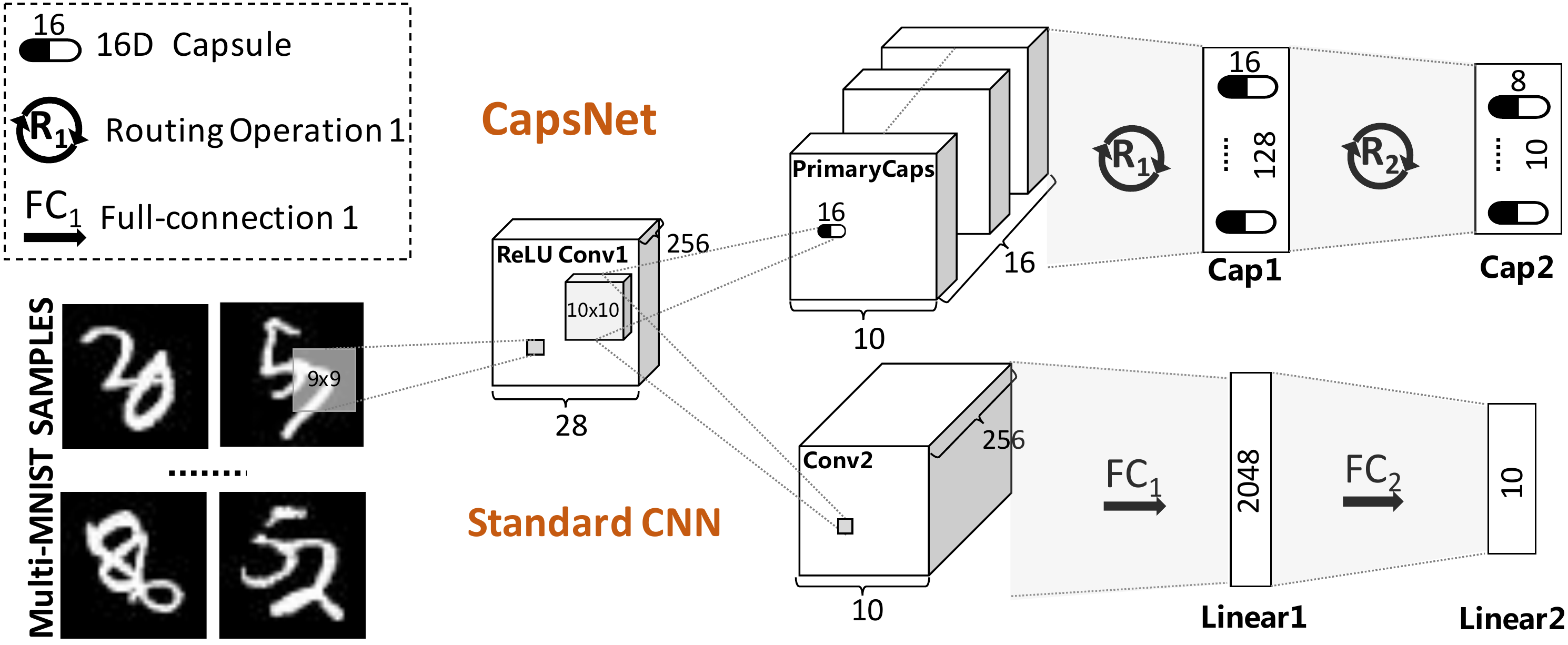}
\par\end{centering}
\caption{\label{fig:cnet-cnn-illu}CapsNet and Standard CNN models. The upper
and lower diagrams show a 4-layer CapsNet as in \cite{Sabour2017}
and a similarly structured standard convolutional neural network,
respectively. Legends in the figure indicate capsule units of certain
dimensions, the routing operation or the fully connection between
layers.}
\end{figure}
\begin{figure}
\begin{centering}
\includegraphics[width=0.9\columnwidth]{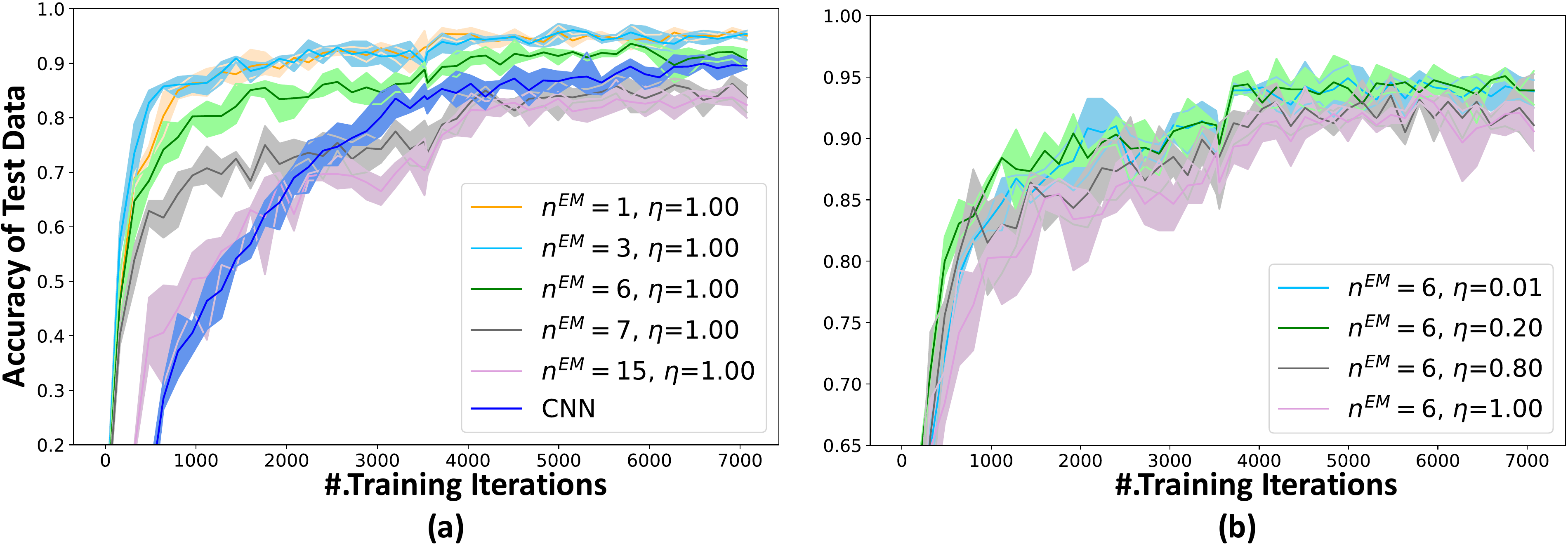}
\par\end{centering}
\caption{\label{fig:n_eta_em}Effects of EM Iteration Number $n^{EM}$ and
Update Rate $\text{\ensuremath{\eta}}$ on CapsNet Training. This
figure shows the training progress in terms of prediction accuracy
on the test dataset against the training steps in mini-batches. Solid
curve represents the mean accuracy in 5 trails using the same $n^{EM}$
and $\text{\ensuremath{\eta}}$. Shaded areas represent one standard
variance of the model performance in the trails. (The figures in this
paper are best viewed in colors.)}
\end{figure}
\textbf{Data}: The task for the models is to recognize two handwritten
digits in one image \cite{Sabour2017}. Each data sample is a $36\times36$
grey-scale image by superposition of two hand-written digit images
from the MNIST dataset \cite{LeCun1998}, with a duplex label of the
two digits. The dataset contains $30,889$ training samples and $4,738$
test samples. Fig. \ref{fig:cnet-cnn-illu} shows a few example images.

\textbf{\textcolor{black}{Models}}: Fig. \ref{fig:cnet-cnn-illu}
illustrates the structure of the CapsNet and standard CNN in the experiment.
The CapsNet is similar to that in \cite{Sabour2017}. A convolutional
layer receives the input and is followed by 3 layers capsules. The
last capsule layer represents the prediction of classes. As a baseline
to assess the training, we have also constructed a conventional neural
network with standard convolutional and linear layers (\emph{i.e.}
standard CNN, as shown in Fig. \ref{fig:cnet-cnn-illu}). Standard
CNN has the same structure as the CapsNet. In each intermediate layer,
we keep the number of total neurons in standard CNN and CapsNet the
same.

\textbf{\textcolor{black}{Experiment}}: We have tested different
\emph{EM iteration numbers} and \emph{EM update rates}, then check
the influence on CapsNet training. Both settings refer to the implementation
of (\ref{eq:cij}). Iteration number $n^{EM}$ indicates the EM loops
which compute the coefficients. The update rate is a parameter we
introduce to stabilize the training - instead of using the original
computation in (\ref{eq:cij}), we use a soft update rule letting
\begin{align}
c_{ij}^{New} & \leftarrow\eta\hat{c}_{ij}+(1-\eta)c_{ij}^{Old}\label{eq:eta}
\end{align}
where $\hat{c}_{ij}$ is the coefficient computed using the original
EM update rule (\ref{eq:cij}), and $\eta$ is the EM update rate.
Testing on various $n^{EM}$ and $\eta$, we evaluate the models being
trained periodically to gauge the progress as well as quality of training.
The evaluation protocol follows that in \cite{Sabour2017}, \emph{i.e.}
a correct prediction requires the model to output the identities of
both digits correctly. All training processes share the same optimization
settings.

Fig. \ref{fig:n_eta_em} shows the model performance during training
under different $n^{EM}$ settings. \mbox{Fig. \ref{fig:n_eta_em}(a)}
shows the process under different $\eta$ settings. We tested each
setting of EM in 5 trails and reported the mean and variance of the
model performance during the training processes. It is noteworthy
that the training of CapsNet saturates with increasing EM iterations
after $n^{EM}$ reaches a small number. In fact, excessive EM iterations
(e.g. $n^{EM}\geq7$ in Fig. \ref{fig:n_eta_em}(a)) impacts the effectiveness
of the training and deteriorates the performance. A possible explanation
is as follows. At the beginning of the training stage, the network
weights have not been conditioned to represent meaningful image elements.
The routing produced by EM is mostly random. We can deprive high-layer
capsules the chance of being exposed to all data samples by forcing
the high-layer capsules to focus only on a small subset of low-layer
inputs. In early stage, those subsets tend to be randomly assorted
without any semantic significance, and the selective representation
scheme (low-high layer association with a sparse matrix $\mathbf{C}:\{c_{ij}\}$)
is more likely to impair rather than to improve the capsules\textquoteright{}
ability to discover meaningful attributes of the entities.

In fact, the observation of the \textquotedblleft early over-routing\textquotedblright{}
phenomenon in Fig. \ref{fig:n_eta_em}(a) has motivated our introduction
of soft EM update scheme as in (\ref{eq:eta}). Fig. \ref{fig:n_eta_em}(b)
shows the model training of $n^{EM}=6$ under different $\eta$-values.
The results indicate the advantage of soft EM updates, e.g. by comparing
the curve for $\eta=0.01$ and that for $\eta=1.0$. 

From the viewpoint of a low-layer capsule $i$, the corresponding
association coefficients $\{c_{i1},c_{i2},\dots\}$ of the capsules
in the layer above can be considered as a probability distribution
over the high-layer capsules. Let $P_{i}^{c}(j)=c_{ij}$ represent
the event that ``entity $i$'s presence is interpreted by the presence
of high-level entity $j$''. In a sense, the entropy of the distribution,
$H[P_{i}^{c}]$, measures the uncertainty at this step in the simulated
cognition process, while forming high-level concepts using low-level
information.

\begin{figure}
\begin{centering}
\includegraphics[width=0.4\columnwidth]{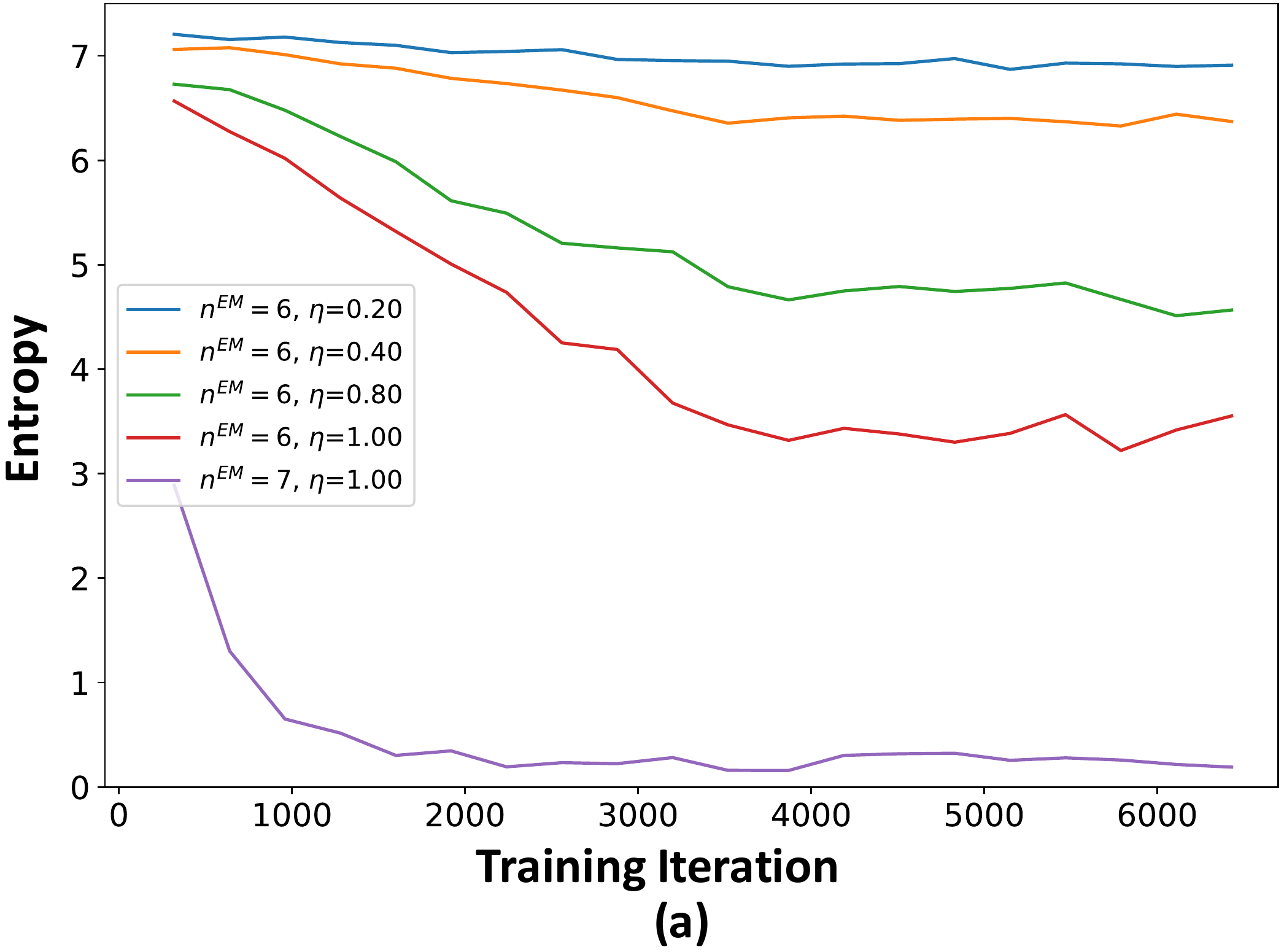} \includegraphics[width=0.4\columnwidth]{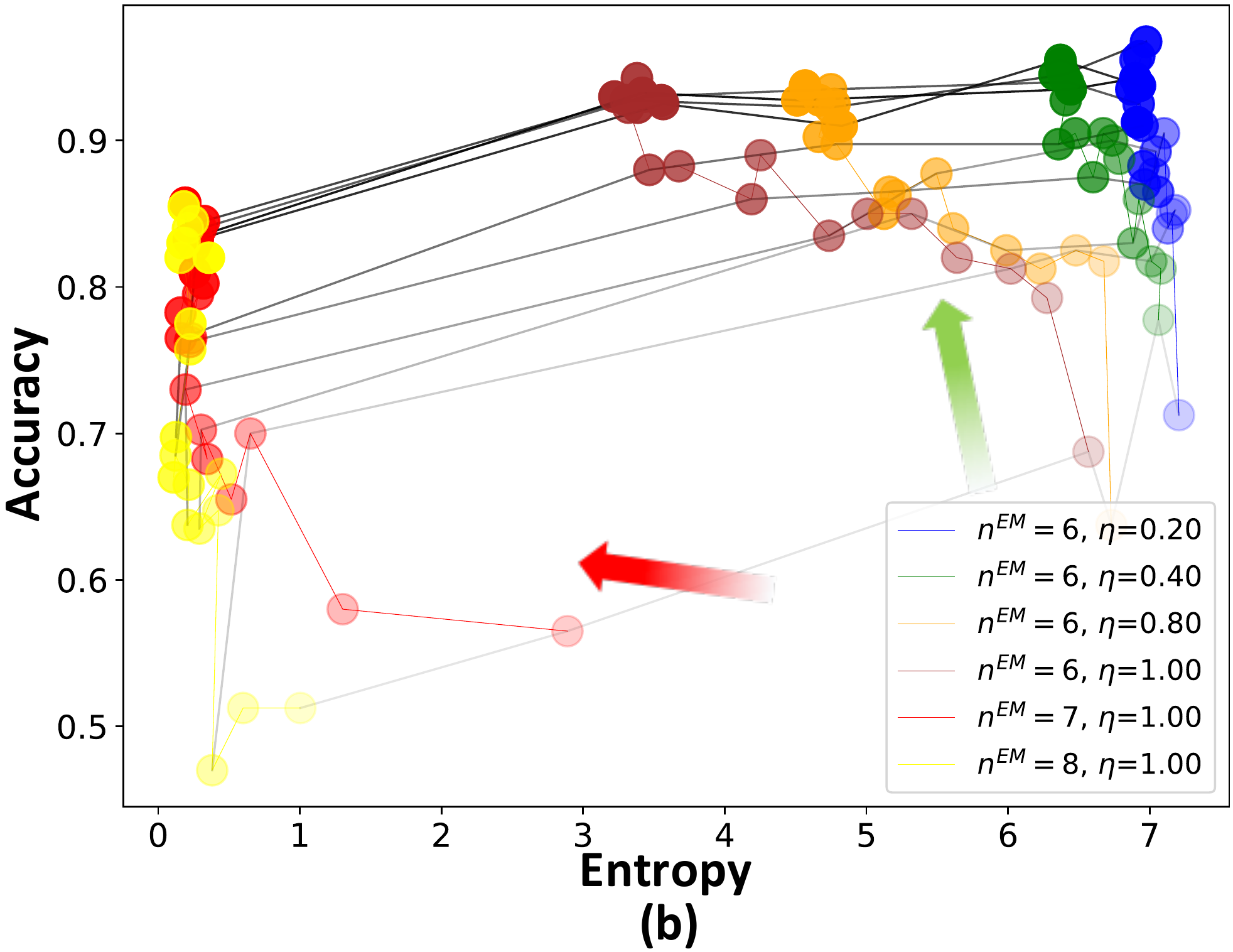}
\par\end{centering}
\caption{\label{fig:entropy}Trends of entropy of the stochastic association
between successive capsule layers during training. \textbf{(a)} shows
the change of entropy along with training iterations, the top four
lines are under different \textgreek{h}-values of fixed EM-iteration
6. Last line shows one \textgreek{h}-value setting of fixed EM-iteration
7. \textbf{(b)} illustrates the trend in the change of entropy and
classification accuracy. Curve marker colors represent different EM-settings.
Transparency indicates training progress. Markers in one gray line
represent models undergone the same number of training iterations.
Red and green arrows show appropriate and excessive reduce of associate
entropy, respectively. See text for more details.}
\end{figure}
Fig. \ref{fig:entropy} shows the trends of the average association
entropy over the training process. In particular, the association
coefficients are from the first EM routing operator $\mathcal{R}_{1}$
as shown in Fig. \ref{fig:cnet-cnn-illu}. Between the two layers,
$\mathcal{R}_{1}$ produces a coefficient matrix $\mathbf{C}^{(n)}$
for each data sample $n$, and the $i$-th row of $\mathbf{C}^{(n)}$
realizes the abovementioned distribution, from which we can compute
an entropy value $H_{i}^{(n)}$. We take the average for the entropy
over every samples $n$ and every low-layer capsules $i$,
\begin{align*}
\bar{H} & =\frac{1}{N\cdot I}\sum_{n=1}^{N}\sum_{i=1}^{I}H[P_{i}^{c,n}]\\
 & =-\frac{1}{N\cdot I\cdot J}\sum_{n=1}^{N}\sum_{i=1}^{I}\sum_{j=1}^{J}c_{ij}^{(n)}\log c_{ij}^{(n)}
\end{align*}
where the superscript $^{n}$ indicates the data sample. The trend
of the average entropy shown in Fig. \ref{fig:entropy}(a) reveals
the mechanism of routing in two aspects:
\begin{enumerate}
\item Reading the plot Fig. \ref{fig:entropy}(a) \emph{vertically}, at
a certain training iteration (x-axis in Fig. \ref{fig:entropy}(a)),
we first compare the average entropy (y-axis in Fig. \ref{fig:entropy}(a))
for different EM settings. As expected, if performing EM with more
stringent updating rules (\emph{i.e.} high updating rate), the less
uncertainty remains in resultant association distribution, and the
entropy reduces.
\item More interestingly, we can also read the plot Fig. \ref{fig:entropy}(a)
\emph{horizontally: }we check how the entropy reached \emph{using}
\emph{a certain EM setting} varies along the model training. The plot
shows that the entropy reduces when the model gets more completely
trained.
\end{enumerate}
\textit{\emph{A possible interpretation of observation 2 is as follows.
When the model fits well to the images, the parts to which individual
capsule units response become clearer and the connections between
layers (coefficients $W$, not to be confused with association $\mathbf{C}$)
become more relevant. Generally speaking, we can be more certain about
whether a low-layer capsule should contribute to the activation of
a high-layer one.}} As an intuitive example, one can usually determine
with more confidence whether a body part belongs to some creature,
and one can with amorphous blob of pixels to some blurry assortment. 

The above understanding leads to a heuristic training strategy: the
EM operation should be modulated such that the certainty of the resultant
association matches the model fitness to the data. In Fig. \ref{fig:entropy}(b),
we plot the average association entropy against the model performance
for models at different training stages and EM settings. Each colored
curve represents one EM operation setting. Each grey line links models
tested after the same number of training steps. The general tendency
in the plot is consistent with our observation in both \mbox{Fig.
\ref{fig:entropy}(a)} and Fig. \ref{fig:n_eta_em}: during training,
model accuracy increases with the association certainty. The green
arrow in Fig. \ref{fig:entropy}(b) intuitively illustrates the phenomenon.
On the other hand, the red arrow in Fig. \ref{fig:entropy}(b), corresponding
to deeply reduced entropy in early training stage, does not bode well
for the training. It indicates that the CapsNet overestimates the
confidence without appropriable fitting, thereby may encounter the
``early over routing'' issue as discussed above.

\subsection{Representation learning\label{subsec:replearn}}

\begin{figure}
\begin{centering}
\includegraphics[height=1.5cm]{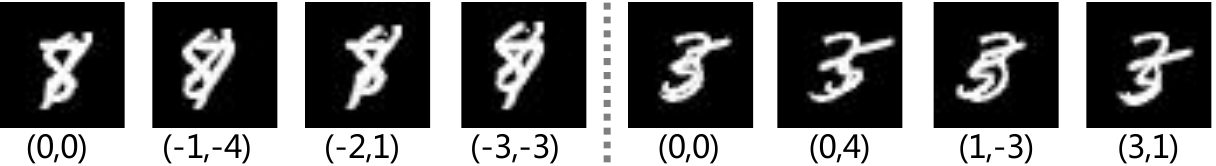}
\par\end{centering}
\caption{\label{fig:rep2images}Example images of 2 shifted and overlapping
hand-written digit images. The left panels shows images of fixed ``7''
and moving ``8''; the right panels are of fixed ``3'' and moving
``5''. The label below each image indicates the offset ($\pm$rows,
$\pm$cols) of the moving part with respect to the fixed part. Take
the (7,8)-images for example, (0,0) means that the image ``8'' is
located with the center overlapping that of ``7'' and (-1,-4) means
that the ``8'' has been shifted 1 units up and 4 units to the left.}
\end{figure}
\begin{figure}
\begin{centering}
\includegraphics[width=0.9\columnwidth]{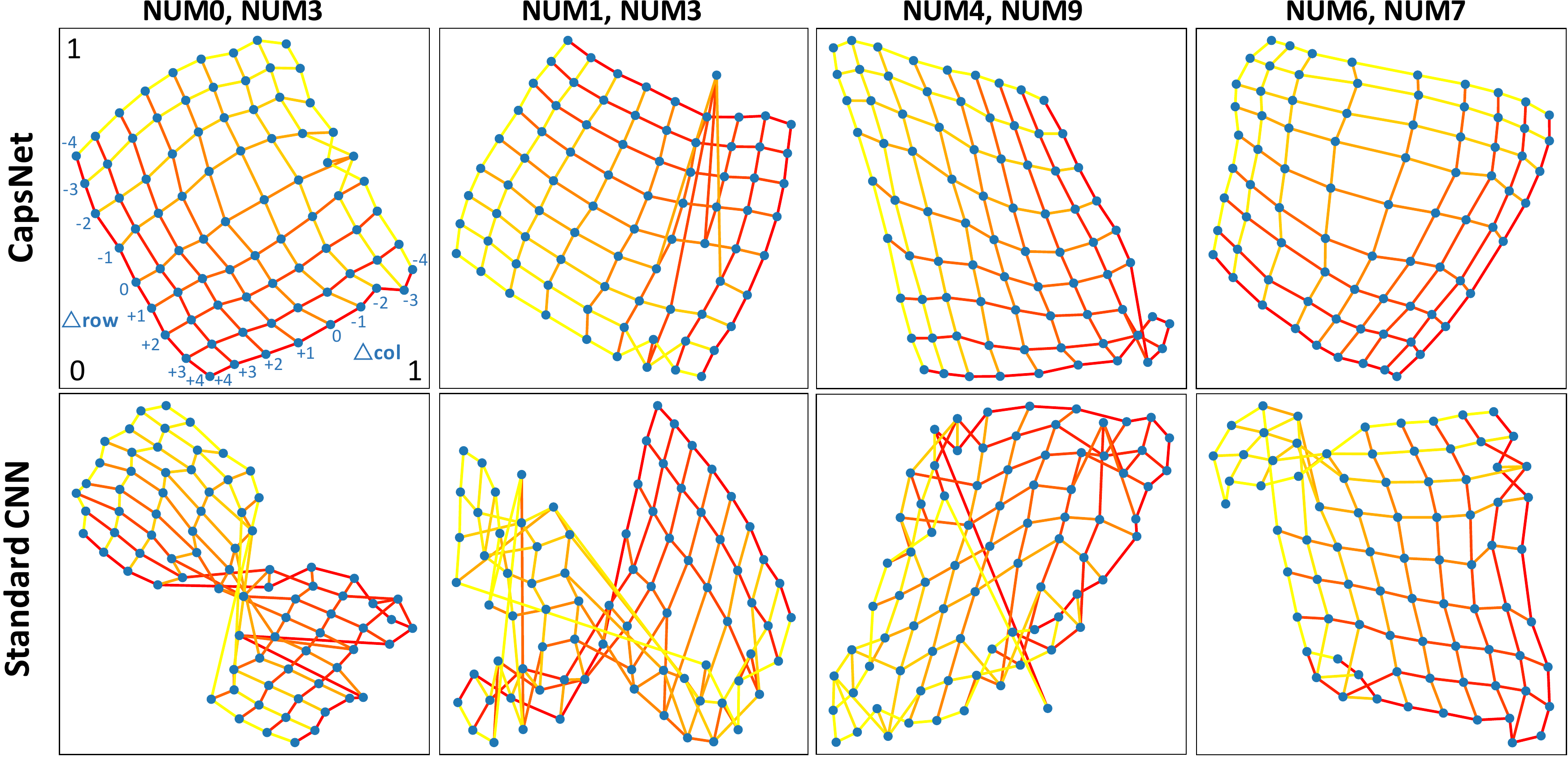}
\par\end{centering}
\caption{\label{fig:Embedding}Embedding of learned representations. This figure
shows the 2D embedding result of CapsNet and Standard CNN on 4 different
test dataset. Points in subplot corresponding to samples in a dataset.
Digits being shifted for the same amount of units and direction have
been connected by a line with certain color. }
\end{figure}
This section presents experiment results on how the CapsNet learned
data representation corresponds to the intrinsic structures in the
population distribution of the data. Beyond assessing the representation
by intuition and visual plausibility, we have especially constructed
a dataset with internal structures induced by geometric transformations.
We perform both quantitative and qualitative assessment of the learned
data representations by CapsNet and standard CNN.

\textbf{Data}: It is common to render data samples encoded by a model
in a 2D plane to examine how well the model has learned to represent
the data \cite{Nie2010,Wang2018,Zhang2015}. In this experiment, we
directly construct test image datasets with known underlying 2D manifold
spanned by geometric translations. The data resemble the multiple
overlapping hand-written digit images. We generate images of two digits
by moving one digit while keeping the other one fixed. The population
of such a dataset is naturally distributed on a 2D manifold. Fig.
\ref{fig:rep2images} shows two example test datasets of digits (fixed-7,
moving-8) and (fixed-3, moving-5), respectively. We shifted the moving
digit horizontally and vertically by $[-4,+4]$ units, resulting in
$9\times9=81$ sample images per test dataset. Note that we discussed
above the \emph{test} datasets. The models are trained on the multi-MNIST
data as in the last experiment. Notably, we evaluate the intermediate
data representations in networks trained for classification, rather
than optimizing the network deliberately for discovering the intrinsic
manifold in the test data.

\textbf{\textcolor{black}{Models}}: In this experiment, we use the
same CapsNet and standard CNN models as described in Subsection \ref{subsec:model-fitting}. 

\textbf{Experiment}: Using trained networks to process the test dataset,
we collect the data representation at an intermediate layer of neurons
/ capsules (the capsule layer after the first routing operation and
the counterpart layer in the standard CNN, see Fig. \ref{fig:cnet-cnn-illu}
for ``Cap1'' and ``Linear1'', respectively). Then we apply manifold
embedding algorithm t-SNE \cite{Maaten2008} to render the learned
representations into $\mathbb{R}^{2}$. Fig. \ref{fig:Embedding}
illustrates the t-SNE $\mathbb{R}^{2}$ embedding of the samples in
several test datasets. The observation is that the data representation
and the corresponding embedding from CapsNet are better aligned with
the internal 2D manifold than those from the standard CNN are. The
quantitative validation is as follows. We computed the Chamfer distance
(CD) \cite{Fan2017} between the regular grid of movements $\left\{ (-4,-4),(-3,-4),\dots,(+4,-4),\dots,(+4,+4)\right\} $
and the $\mathbb{R}^{2}$ embedding. The table below shows the mean
and variance of the Chamfer distance.

\vspace{2pt}{\centering\includegraphics[width=4cm]{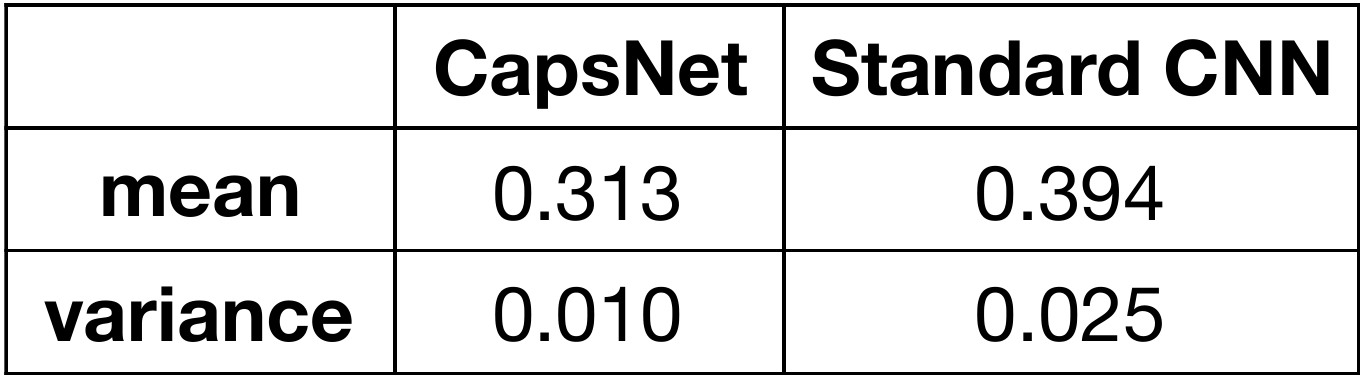}

}

Compared to the intermediate data representation of the standard CNN,
that of CapsNet is more strongly related to the geometrically meaningful
structures of the data population. Recall that the models were trained
as classifiers, which means that the CapsNet discovered relevant data
representation \emph{without} explicit training goal of such structures.
\begin{figure}
\begin{centering}
\includegraphics[width=0.9\columnwidth]{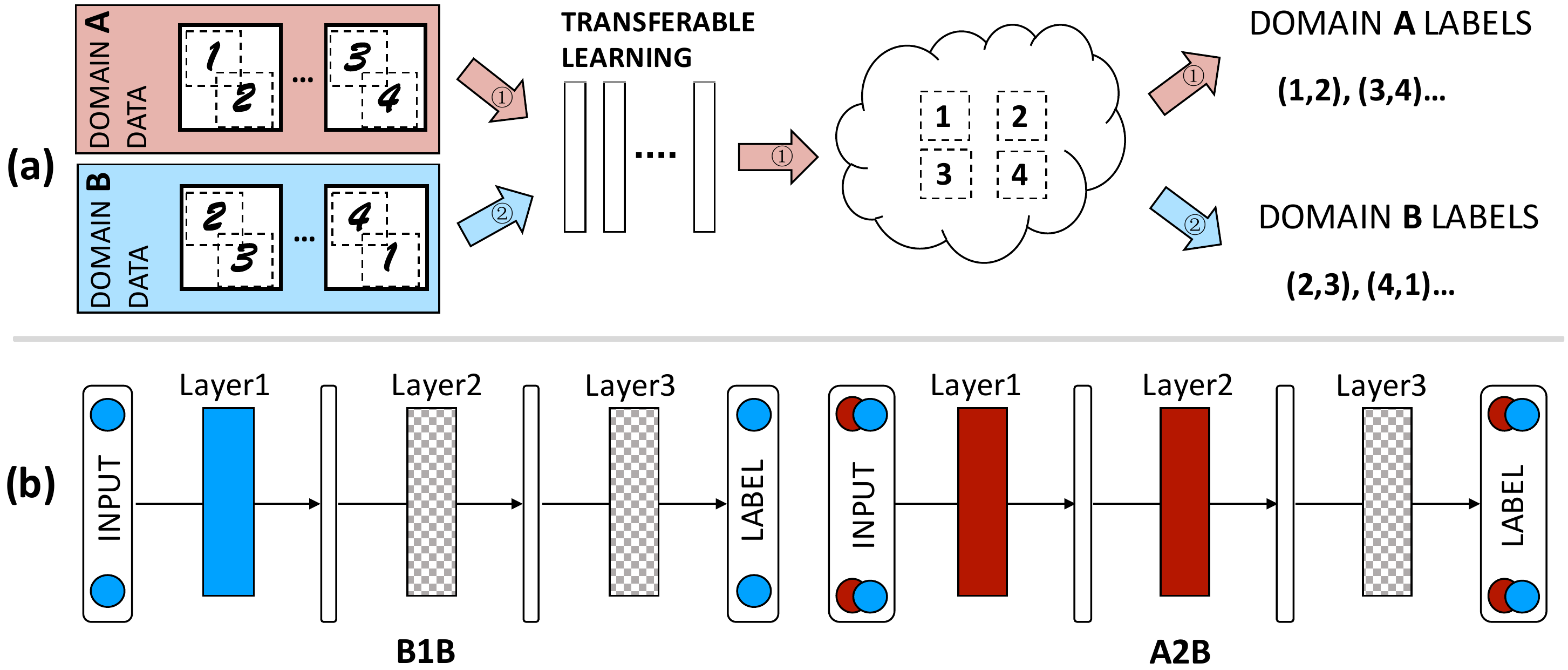}
\par\end{centering}
\caption{\label{fig:transfer_structure}Example Testing Schemes of Domain Transfer.
Colors represent domain, red for A and blue for B. \textbf{(a)} Domain
data are superposition of 2 hand-written digits. Domain A and B contain
different combinations of digits. \textbf{(b)} The tests are to determine
how representations produced by pre-trained layers can help new cross-/same-domain
tasks. Chessboard pattern stands for layers to re-adjust. B1B represents
keeping 1 pre-trained layer on domain B and re-adjust the top 2 layers
also on domain B. A2B represents keeping 2 pre-trained layers on domain
A and re-adjust the top 1 layer on domain B. See text and refer to
\cite{Yosinski2014} for more detailed discussions of the testing
protocol.}
\end{figure}

\subsection{Representation generalisation}

Meaningful data representation can facilitate knowledge transfer or
generalization to distinctive cognition tasks. In this experiment,
we further investigate how a trained CapsNet generalizes beyond the
original task, in particular, how transferrable the CapsNet are. 

\textbf{Data}: We have made two subsets of the multi-MNIST image data
for the test of transfer-domain fitness. The two subsets, namely $S_{A}$
and $S_{B}$, are constructed so that i) each two-digit class of images
are exclusively within $S_{A}$ \emph{or} $S_{B}$, and ii) $S_{A}$
and $S_{B}$ both contain a complete set of digits. The motivation
of using such $S_{A}$ and $S_{B}$ is as follows. Condition i) ensures
that during training, the model does not use the images of the same
two-digit label on which it will be tested, and Condition ii) ensures
that the model has seen all necessary concepts, \emph{i.e.} the individual
digits, in order to successfully perform the new task. For example,
if the model to be tested on images containing digits $(7,8)$, then
we do not use images of $(7,8)$ in the training stage. Instead, the
training data contain the appearance of both 7 and 8 in images such
as $(7,9)$, $(1,8)$, etc. as shown in Fig. \ref{fig:transfer_structure}(a).

\textbf{\textcolor{black}{Models}}: The CapsNet and standard CNN models
are similar to those used in Subsection \ref{subsec:model-fitting}
with one more capsule/CNN layer, respectively. The extra layer is
for testing the transferability of the neurons at different layers.

\textbf{Experiment}: We follow a similar test protocol as that in
\cite{Yosinski2014}: the models are trained in domain $S_{A}$ and
tested in domain $S_{B}$. We re-adjust the last 1 or 2 layers using
training data of domain $S_{B}$, while keeping the remaining net
parameters as trained in $S_{A}$. Such models are called A1B (1 layer
fixed to $S_{A}$-training, 2 layers adjusted on $S_{B}$) or A2B
(2 layers fixed to $S_{A}$-training, 1 layer adjusted on $S_{B}$).
As a control test, the experiment also includes networks prepared
following the above protocol with the difference that the first training
pass is on the target domain $S_{B}$. We call such control set of
models B1B (fixing 1 layer and re-adjusting 2) and B2B (fixing 2,
re-adjusting 1), respectively. Fig. \ref{fig:transfer_structure}(b)
illustrates the test schemes of B1B and A2B.

Keeping lower layers fixed and re-adjusting the higher ones breaks
the coupling between the layers formed during training. BnB schemes
test if the data representation of the lower layers can be decoupled
from the original higher ones. AnB schemes tend to be more challenging,
where the learned representation must survive cross-domain readaption.
Cross-domain readaption can benefit from representation contains intermediate-level
knowledge that is relevant to both tasks, e.g. the appearance of individual
digits in this experiment.

\begin{figure}
\begin{centering}
\includegraphics[width=0.6\columnwidth]{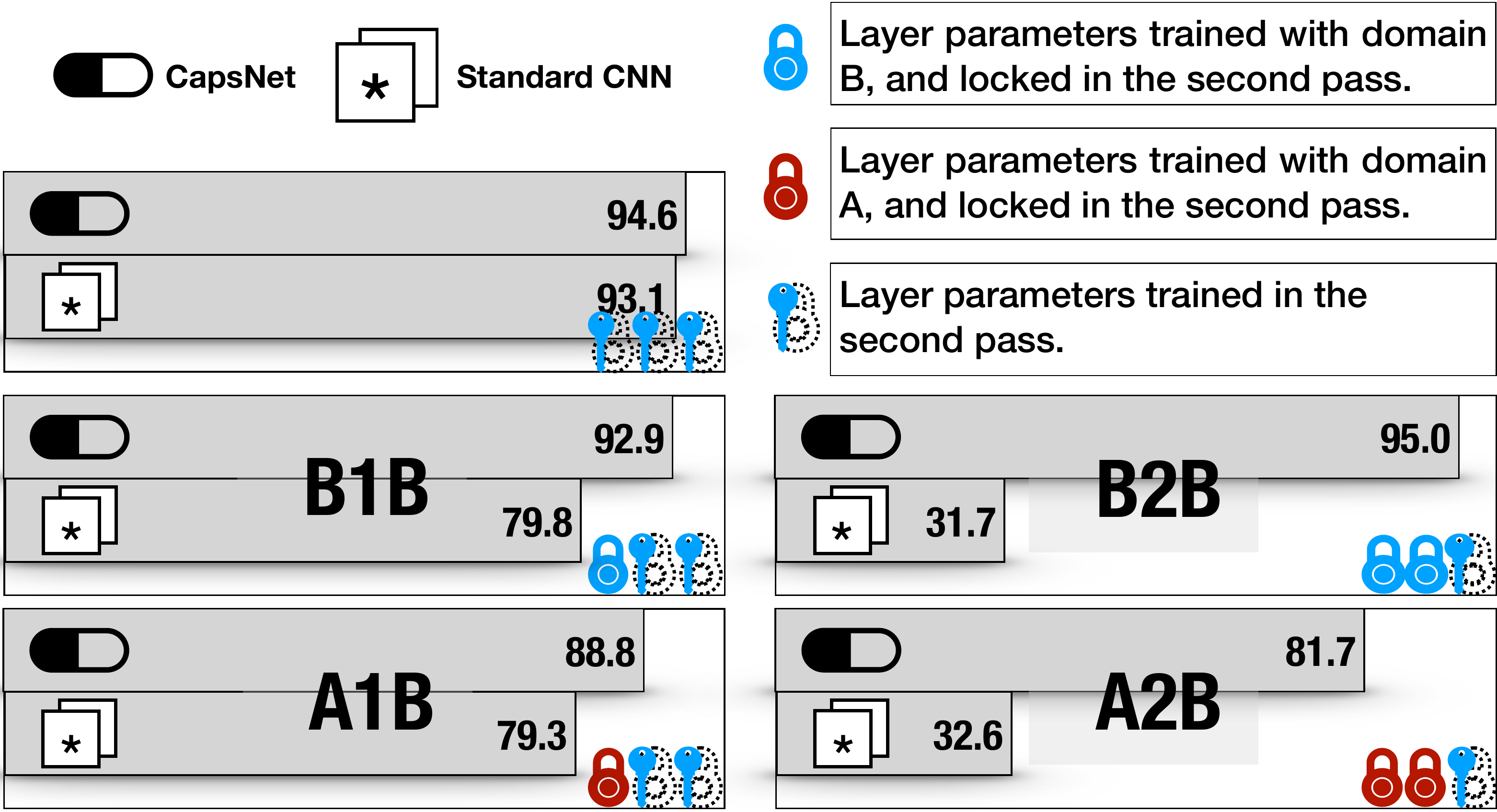}
\par\end{centering}
\caption{\label{fig:result_transfer}Transfer Performance of CapsNet and Standard
CNN. The figure shows model performance on different testing schemes.
The first plot shows the baseline model performance of standard supervised
learning on domain B.}
\end{figure}
Fig. \ref{fig:result_transfer} displays the results of CapsNet and
standard CNN on different transfer test schemes. We have observed
that breaking the coupling between the layers significantly reduced
the fitness of standard CNN, regardless of the domain on which the
original model was trained. On the other hand, CapsNet representation
can be successfully used by newly learned higher layers. When the
new task is cross-domain, CapsNet has a minor performance drop. Nevertheless,
the CapsNet representations remain satisfactorily relevant on AnB
tasks, which supports the claim that the intermediate level capsules
can capture knowledge on appearance of meaningful object parts \cite{Sabour2017}.\textbf{}\vspace{-5pt}

\section{Conclusion\label{sec:con}}

In this paper we investigate several important aspects of CapsNet,
including model learning, attributes of learned data representations
and generality of the representations. Our tests demonstrate that
appropriate routing operation plays a significant role in CapsNet
training. In the early stage of training, the routing between capsules
should contain a level of \emph{uncertainty}; early over-confidence
about the routing tends to impose excessive limits on the training
process, which leads to suboptimal models. CapsNet can produce data
representation with interesting attributes. To explore such attributes,
we especially designed a test using image data on a 2D manifold spanned
by geometric transformations. The test shows that compared to standard
CNN, CapsNet can capture more faithfully the global manifold structures
in data. Moreover, following test protocol in \cite{Yosinski2014},
we show the representation by CapsNet is more transferrable than that
by standard CNN.

\bibliographystyle{plain}
\bibliography{biblo}

\end{document}